\documentclass[conference]{IEEEtran}
\IEEEoverridecommandlockouts
\usepackage{cite}
\usepackage{amsmath,amssymb,amsfonts}
\usepackage{algorithmic}
\usepackage{graphicx}
\usepackage{textcomp}
\usepackage{xcolor}
\usepackage{multirow}
\def\BibTeX{{\rm B\kern-.05em{\sc i\kern-.025em b}\kern-.08em
    T\kern-.1667em\lower.7ex\hbox{E}\kern-.125emX}}
\begin{document}

\title{Toward Robust EEG‑based Intention Decoding during Misarticulated Speech in Dysarthria\\
% {\footnotesize \textsuperscript{*}Note: Sub-titles are not captured for https://ieeexplore.ieee.org  and
% should not be used}
\thanks{This work was partly supported by Institute of Information \& Communications Technology Planning \& Evaluation (IITP) grant funded by the Korea government (MSIT) (No. RS-2019-II190079, Artificial Intelligence Graduate School Program (Korea University), No. RS-2021-II-212068, Artificial Intelligence Innovation Hub, and No. RS-2024-00336673, AI Technology for Interactive Communication of Language Impaired Individuals).}
}

\author{

\IEEEauthorblockN{Ha-Na Jo}
\IEEEauthorblockA{\textit{Dept. of Artificial Intelligence} \\
\textit{Korea University} \\
Seoul, Republic of Korea \\ 
hn\_jo@korea.ac.kr}

\and

\IEEEauthorblockN{Jung-Sun Lee}
\IEEEauthorblockA{\textit{Dept. of Artificial Intelligence} \\
\textit{Korea University} \\
Seoul, Republic of Korea \\ 
jungsun\_lee@korea.ac.kr}

\and

\IEEEauthorblockN{Eunyeong Ko}
\IEEEauthorblockA{\textit{Dept. of Artificial Intelligence} \\
\textit{Korea University} \\
Seoul, Republic of Korea \\
eunyeong\_ko@korea.ac.kr} \\
}

\maketitle

% 오타 검수
% 단위 앞에 띄어쓰기
% 약어 나오는 타이밍
% ref 제목 통일, iso4 약어로 수정, 페이지 통일
% fig, table 제목 수정

\begin{abstract}
Dysarthria impairs motor control of speech, often resulting in reduced intelligibility and frequent misarticulations. Although interest in brain–computer interface technologies is growing, electroencephalogram (EEG)-based communication support for individuals with dysarthria remains limited. To address this gap, we recorded EEG data from one participant with dysarthria during a Korean automatic speech task and labeled each trial as correct or misarticulated. Spectral analysis revealed that misarticulated trials exhibited elevated frontal–central delta and alpha power, along with reduced temporal gamma activity. Building on these observations, we developed a soft multitask learning framework designed to suppress these nonspecific spectral responses and incorporated a maximum mean discrepancy–based alignment module to enhance class discrimination while minimizing domain-related variability. The proposed model achieved F1-scores of 52.7 \% for correct and 41.4 \% for misarticulated trials—an improvement of 2 \% and 11 \% over the baseline—demonstrating more stable intention decoding even under articulation errors. These results highlight the potential of EEG-based assistive systems for communication in language impaired individuals.
\end{abstract}


\section{INTRODUCTION}
% Expressive aphasia is a neurological disorder characterized by impaired speech production despite preserved comprehension \cite{aphasia-review}. Patients often produce misarticulated speech that fails to convey their intended meaning, leading to significant communication barriers in daily and social interactions. This mismatch between intention and articulation hinders effective communication, often causing frustration, social withdrawal, and reduced quality of life. Therefore, decoding patients’ true communicative intentions beyond their verbal output is crucial for supporting meaningful interaction and rehabilitation.

Dysarthria is a motor speech disorder characterized by impaired control of the muscles responsible for articulation, phonation, and prosody, despite preserved language comprehension \cite{dysarthria-review}. Patients often produce distorted or unintelligible speech that fails to reflect their intended message, creating substantial communication barriers in everyday and social interactions. This discrepancy between intended speech and its articulatory realization impedes effective communication and may lead to frustration, reduced social participation, and diminished quality of life. Consequently, decoding a patient’s communicative intention beyond the accuracy of their spoken output is essential for supporting meaningful interaction and clinical rehabilitation.

To address these communication challenges, brain–computer interface (BCI) technology provides a direct link between neural activity and external devices without requiring overt speech or motor output using machine learning or artificial intelligence models \cite{prml-neural}. BCIs have been extensively applied in domains such as cognitive workload monitoring, motor imagery, and emotional state recognition, offering alternative communication channels for individuals with severe speech or motor impairments \cite{cognitive, prml-motor, prml-emontion}. These advances suggest that BCIs could be leveraged to decode communicative intention in language impaired individuals by interpreting neural correlates of attempted or misarticulated speech.

Furthermore, recent studies have investigated speech-related decoding across multiple speech modes—including overt, imagined, perceived, mimed, and whispered speech—aiming to infer linguistic content such as words, sentences, or semantic categories directly from brain signals \cite{imagined, jslee}. While these studies demonstrate promising progress in speech decoding, they have primarily focused on healthy participants under highly controlled laboratory conditions, where subjects silently imagine speech without overt articulation \cite{jslee, prml-changes}. Consequently, the generalizability of such systems to real-world or clinical contexts, especially in patients with impaired articulation, remains limited \cite{realworld}.

\begin{figure*}[thpb]
  \centering
  \includegraphics[width=\textwidth]{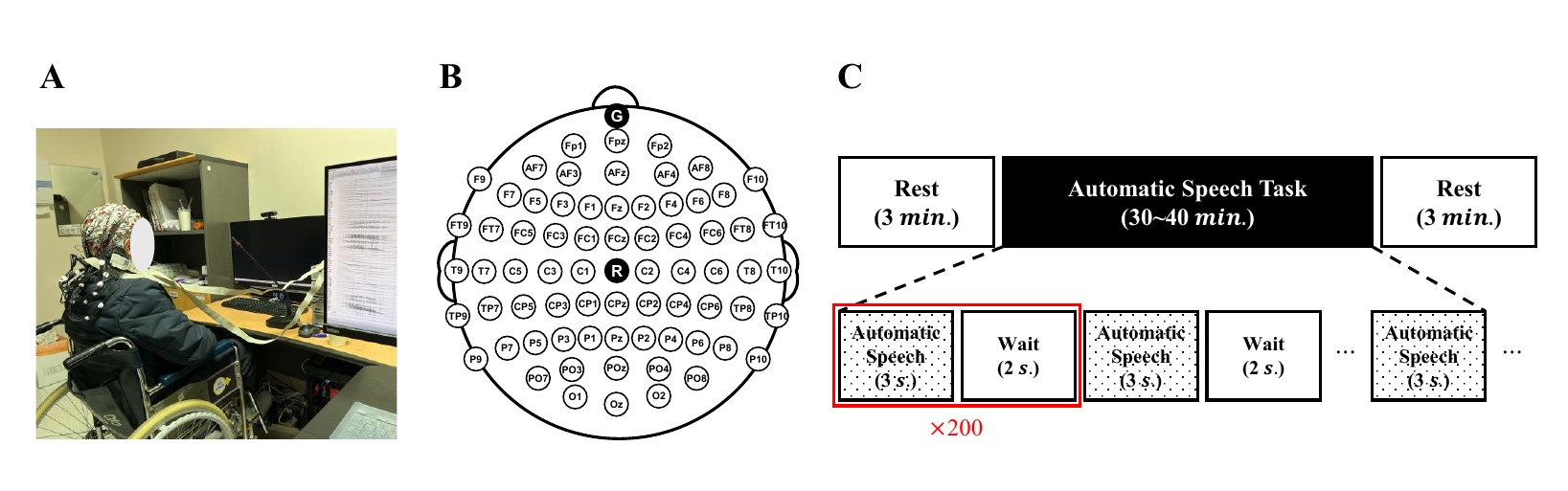}
  \caption{Experimental settings. A) Experimental environment, B) Electrode montage used in the experiment, showing the placement of electrodes according to the international 10–20 system (G = ground, R = reference), C) Experiment paradigm, the automatic speech consisted of four different utterances, which were presented 50 times each in a randomized order, for a total of 200 trials.}
  \label{figure1}
\end{figure*}

\begin{figure}[thpb]
  \centering
  \includegraphics[width=\linewidth]{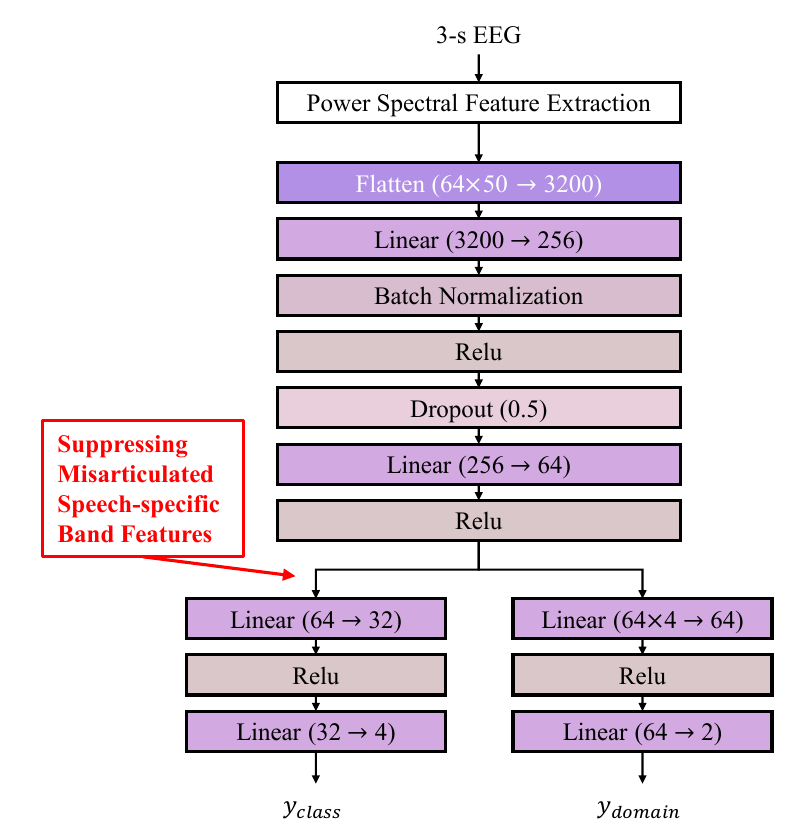}
  \caption{Architecture of the proposed soft multitask learning model. Power spectral features extracted from 3-second EEG segments are processed through shared fully connected layers, followed by two task-specific branches for speech-class classification (\(y_{class}\)) and domain discrimination (\(y_{domain}\)).}
  \label{figure2}
\end{figure}

In this study, we designed a paradigm based on Korean automatic speech tasks that can be performed by individuals with dysarthria. During the task, both correct and misarticulated trials were recorded from a patient, and spectral analyses revealed elevated delta and alpha activity over frontal–central regions, along with reduced gamma power in temporal areas during misarticulated trials \cite{prml-continuous}. Building on these observations, we propose a soft multitask learning model with maximum mean discrepancy (MMD) regularization to robustly decode patients’ communicative intentions regardless of articulation quality \cite{softmulti, mmd}. In particular, the class branch is designed to de-emphasize spectral features in the delta, alpha, and gamma bands, each of which showed statistically significant differences between trials. The proposed architecture follows the design philosophy of compact and task-specific neural networks \cite{prml-multilayer, prml-integrated}, emphasizing structural efficiency and adaptive feature extraction in limited-data environments. This approach aims to achieve intention-level decoding, providing a step toward BCI communication systems for patients.

This study offers the following contributions:

\begin{enumerate}
    \item We introduce a novel experimental paradigm utilizing automatic speech tasks that allow language impaired patients to produce both correct and misarticulated speech within a controlled yet naturalistic setting.
    \item Through spectral analyses of EEG signals, we identified distinct neurophysiological patterns during misarticulated trials: increased frontal–central delta activity associated with compensatory motor effort, elevated alpha power reflecting inhibitory control, and reduced temporal gamma activity indicating weakened sensory–motor integration.
    \item To address the decoding degradation caused by these articulation-specific neural fluctuations, we propose a soft multitask learning model with MMD regularization that attenuates statistically significant alpha and gamma band features from class learning, thereby enabling robust intention decoding across both correct and misarticulated speech and achieving notable improvements over baseline models.
\end{enumerate}

\section{METHODS}
\subsection{Experimental Design}
A 49-year-old male participant diagnosed with ataxic type of dysarthria was recruited for this study. His baseline speech intelligibility was evaluated between 61 \% and 80 \%, indicating moderate impairment in articulation. Prior to participation, ethical approval was obtained from the Institutional Review Board of Korea University (KUIRB-2023-0429-01).

To investigate neural patterns underlying correct and misarticulated speech, we designed a Korean automatic speech paradigm. Four familiar Korean automatic word sequences were used: “ga-na-da-ra-ma,” “wol-hwa-su-mok-geum,” “hana-dul-set-net-daseot,” and “a-e-i-o-u,” each representing overlearned phonemic patterns similar to English alphabet or counting sequences \cite{automatic}. In each trial, the last syllable of a sequence was omitted and visually presented on a screen, prompting the participant to verbally complete it within 3 s. following the common paradigm \cite{prml-subject}. A 2 s. rest interval followed each trial. Each sequence consisted of 50 random repetitions, resulting in 200 trials in total. Trials were labeled as correct if the response matched the target within the time limit, or as misarticulated if the response was incorrect or absent.

\begin{figure*}[thpb]
  \centering
  \includegraphics[width=\textwidth]{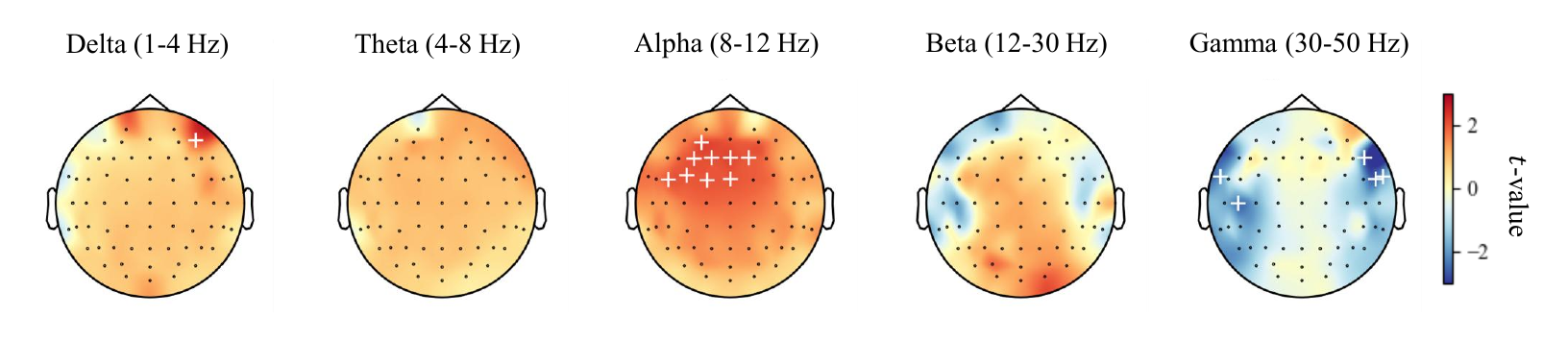}
  \caption{Topographical maps of \textit{t}-values for the comparison of power spectral density between correct and misarticulated sessions across five frequency bands (delta, theta, alpha, beta, and gamma). Positive \textit{t}-values represent higher power in misarticulated trials, while negative \textit{t}-values represent lower power compared to correct trials. ‘+’ denotes statistical significance (\textit{p} $<$ 0.05, FDR corrected).}
  \label{figure3}
\end{figure*}

\subsection{Data Acquisition}

EEG data were recorded using a 64-channel BrainVision system arranged according to the international 10--20 montage \cite{eegplacement, prml-eegclf}. The reference electrode was placed at Cz. Signals were sampled at 500 Hz. A band-pass filter between 1 Hz and 50 Hz was applied to remove slow drifts and high-frequency noise \cite{prml-csp}. Electrode impedance was maintained below 20 k$\Omega$ throughout the experiment \cite{impedence}. All recordings were conducted in a quiet, partitioned testing area to minimize external noise, and visual stimuli were presented on a monitor synchronized with EEG acquisition via trigger signals (Fig. \ref{figure1}).

% \begin{table}[t]
% \centering
% \caption{Classification accuracy (\%) across models for all, corrected, and misarticulated trials.}
% \resizebox{\columnwidth}{!}{%
% \begin{tabular}{lccc}
% \hline
% \textbf{Model} & \textbf{All} & \textbf{Corrected} & \textbf{Misarticulated} \\
% \hline
% Baseline        & 40.9 & 44.8 & 32.0 \\
% \textbf{Ours} & \textbf{48.1} & \textbf{48.3} & \textbf{48.0} \\
% \hline
% \end{tabular}}
% \label{table1}
% \end{table}

\begin{table}[]
\centering
\caption{Classification accuracy (\%) and F1-score across models for all, correct, and misarticulated trials.}
\resizebox{\columnwidth}{!}{%
\begin{tabular}{lcccc}
\hline
\multirow{2}{*}{\textbf{Model}} & \multirow{2}{*}{\textbf{Accuracy}} & \multicolumn{3}{c}{\textbf{F1-score}}                      \\ \cline{3-5} 
                                &                                    & \textbf{All}  & \textbf{Correct} & \textbf{Misarticulated} \\ \hline
Baseline                        & 40.9                               & 40.3          & 50.4             & 30.0                    \\
\textbf{Ours}                   & \textbf{45.5}                      & \textbf{44.7} & \textbf{52.8}    & \textbf{41.4}           \\ \hline
\end{tabular}}
\label{table1}
\end{table}

\subsection{Spectral Analysis} 

To investigate frequency-specific characteristics associated with correct and misarticulated speech, we performed power spectral density (PSD) analysis on the EEG signals \cite{my}. The PSD was estimated using the Fast Fourier Transform method, which efficiently computes the spectral power distribution with low computational cost \cite{prml-multiresolution, fft}. 

% The PSD for a signal \( x(t) \) was estimated using the Welch method, which averages the squared magnitude of the Fourier-transformed signal across overlapping windows:
% \begin{equation}
% P_{xx}(f) = \frac{1}{N} \sum_{i=1}^{N} \frac{1}{U} \left| \mathcal{F}\{w(t)x_i(t)\} \right|^2,  
% \end{equation}
% where \( w(t) \) denotes the window function, \( \mathcal{F}\{\cdot\} \) is the Fourier transform, and \( U \) is a normalization factor. 

For each EEG channel and frequency band, PSD values were estimated using the Welch method. All statistical results were corrected for multiple comparisons using the false discovery rate (FDR) with the Benjamini–Hochberg procedure, and significance was determined at an alpha level of \textit{p} $<$ 0.05.

% and statistically compared between correct and misarticulated trials using independent-sample \( t \)-tests. Statistical significance was determined at a threshold of \( p < 0.05 \). 

\subsection{Proposed Model and Training Strategy}

To achieve robust intention decoding under misarticulated speech, we proposed a tiny soft multitask learning framework that jointly optimizes speech-class prediction and domain (speech accuracy) discrimination, inspired by compact architectures \cite{prml-pattern, prml-biologically}. The overall architecture is shown in Fig.~\ref{figure2}.

EEG data with dimensions $(C \times F)$, where $C$ and $F$ denote the number of channels and frequency bins, were input to a shared encoder that extracts latent spatial–spectral features. The encoder output was branched into two predictors: (1) a \textit{class predictor} for utterance classification and (2) a \textit{domain predictor} for distinguishing correct from misarticulated trials. The class branch suppresses the delta, alpha and gamma range that showed significant spectral differences to promote class-discriminative yet domain-invariant representations.

Finally, MMD regularization term was applied between the encoded representations of correct and misarticulated trials. The total loss was defined as:
\begin{equation}
\mathcal{L}_{total} = \mathcal{L}_{class} + \lambda_1 \mathcal{L}_{domain} + \lambda_2 \mathcal{L}_{MMD},
\end{equation}
where $\mathcal{L}_{class}$ and $\mathcal{L}_{domain}$ denote the class and domain losses, and $\mathcal{L}_{MMD}$ measures the discrepancy between the embeddings of correct and misarticulated trials. The coefficients $\lambda_1$ and $\lambda_2$ were empirically set to 0.3 and 0.3, respectively.

\section{RESULTS AND DISCUSSION}

\subsection{Spectral Characteristics of Misarticulated Speech}

% The power spectral analysis revealed distinct frequency-dependent modulations between correct and misarticulated trials. As shown in Fig. \ref{figure3}, significant power increases were observed in the low-frequency range (1–12 Hz) during misarticulated speech, particularly within the delta (1–4 Hz), theta (4–8 Hz), and alpha (8–12 Hz) bands. Delta activity was globally enhanced across most cortical regions, while theta–alpha enhancement was primarily localized over the frontal channels. 

The power spectral analysis revealed two distinct frequency-specific modulations during misarticulated trials. As shown in Fig. \ref{figure3}, delta and alpha band power exhibited a pronounced increase over frontal–central regions, whereas gamma band activity showed a marked reduction localized to the temporal channels.

These frequency patterns align with neurophysiological features reported in dysarthria. Elevated frontal–central delta activity is commonly observed when greater compensatory motor demands are present, while increased alpha power has been associated with heightened inhibitory control during effortful or unstable speech production \cite{delta2, alpha}. In contrast, reductions in temporal gamma activity are typically linked to weakened sensory–motor integration, particularly when auditory–motor coordination becomes less reliable \cite{temporal-gamma}. Taken together, these oscillatory modulations indicate that misarticulated speech in dysarthria is accompanied by neural signatures reflecting increased motor effort, enhanced inhibitory engagement, and reduced sensory–motor coupling. Such articulation-driven variability presents challenges for conventional decoding models, underscoring the need for approaches that remain robust to these fluctuations.

\subsection{Performance of the Proposed Learning Framework}

% To evaluate the proposed model, three configurations were tested: (1) a baseline encoder-only model, (2) with soft multitask learning, and (3) with both multitask learning and MMD regularization. The dataset (72 \% correct, 28 \% misarticulated) was split 8:2 into training and testing sets while maintaining class balance. As shown in Table~\ref{table1}, the baseline achieved 41.4 \% accuracy for correct trials but failed to generalize to misarticulated trials (0.0 \%). Adding soft multitask learning improved robustness to 48.4 \% and 27.3 \% for correct and misarticulated trials, respectively, while the inclusion of MMD regularization further enhanced performance to 58.6 \% and 45.5 \%.

To evaluate model performance, we compared a baseline encoder-only model with our multitask framework incorporating MMD regularization. As shown in Table~\ref{table1}, the baseline achieved 40.9 \% accuracy and poorly generalized to misarticulated trials (F1-score: 30.0 \%). In contrast, our model improved overall accuracy to 45.5 \% and achieved higher F1-scores across all categories, including 52.8 \% for correct and 41.4 \% for misarticulated trials, demonstrating greater robustness to articulation-related variability. 

These results demonstrate that the proposed model effectively mitigates articulation-specific overfitting by jointly learning class and domain information through a shared encoder. By softly sharing parameters between the two branches, the network is encouraged to detect domain-related variations while simultaneously refining class-discriminative features. Importantly, the class branch is trained to suppress the nonspecific spectral patterns that characterize misarticulated trials, allowing the model to prioritize neural activity that more faithfully reflects the participant’s intended speech content rather than the quality of motor execution. The improved generalization to misarticulated trials indicates that the model captures intention-relevant neural representations that remain stable even when articulation fails, highlighting its ability to down-weight articulation-driven variability and emphasize the underlying cognitive intent.

% Although the overall accuracy for correct trials remained moderate compared to healthy controls, this reflects the participant’s baseline speech intelligibility (61–80 \%) and the inherent variability in aphasic cortical dynamics \cite{aphasia-eeg}. Nevertheless, the model’s robustness across articulation states highlights its potential for practical application in patient-centered BCIs. By decoupling intention decoding from motor output quality, this approach could enable reliable communication even when overt speech is impaired or inconsistent. Future extensions with larger participant groups and adaptive transfer learning strategies may further enhance its generalizability, paving the way toward clinical integration of intention-based neural decoding for expressive aphasia.

\section{CONCLUSIONS}

This study demonstrates that communicative intent can be robustly decoded from EEG signals even when overt speech is distorted, revealing articulation-invariant neural representations in dysarthria. These findings suggest a promising direction for developing BCI systems that emphasize intention-level decoding rather than motor output, offering a foundation for restoring reliable communication in individuals with dysarthria. In future work, integrating multimodal approaches such as vision-based cues or reinforcement learning–driven adaptation could further extend this framework toward real-world, context-aware intention decoding and assistive communication systems \cite{prml-motion, prml-deep}.

% \section*{References}
\bibliographystyle{IEEEtran}
\bibliography{ref}

\end{document}